\documentclass{article}

\usepackage[preprint]{neurips_2025}

\usepackage[utf8]{inputenc} 
\usepackage[T1]{fontenc}    
\usepackage{hyperref}       
\usepackage{url}            
\usepackage{booktabs}       
\usepackage{amsfonts}       
\usepackage{nicefrac}       
\usepackage{microtype}      
\usepackage{xcolor}         
\usepackage{multirow}
\usepackage{tabularx}
\usepackage{threeparttable}
\usepackage{enumitem}
\usepackage{nicefrac}       
\usepackage{microtype}      
\usepackage{xcolor}         
\usepackage{multirow}
\usepackage{tabularx}
\usepackage{threeparttable}
\usepackage{enumitem}
\usepackage{algorithm}
\usepackage{setspace}
\usepackage{algorithmicx}
\usepackage{latexsym}
\usepackage{graphicx}
\usepackage{amsmath}
\usepackage{algorithm}
\usepackage{algorithmicx}
\usepackage{algpseudocode}
\usepackage{multirow}
\usepackage{tabularx}
\usepackage{algorithm}
\usepackage{setspace}
\usepackage{algorithmicx}
\usepackage{amsfonts}  
\usepackage{threeparttable}
\usepackage{enumitem}
\usepackage{amsmath}
\usepackage{graphicx}
\usepackage{multirow}
\usepackage{threeparttable}
\usepackage{tabularx}
\usepackage{algorithm}
\usepackage{setspace}
\usepackage{hyperref} 
\usepackage{xcolor}
\usepackage{soul}
\usepackage{cleveref}
\usepackage{colortbl}
\usepackage{fontawesome5}
\usepackage{tikz}
\usetikzlibrary{shapes.geometric, arrows, positioning, trees, mindmap}
\usetikzlibrary{matrix, positioning, arrows.meta}

\usepackage{forest}
\usepackage{geometry}
\usepackage{makecell}   
\usepackage{amssymb}    
\usepackage{pifont}     

\usepackage[most]{tcolorbox}

\newcommand{\secref}[1]{\hyperref[#1]{\S\ref*{#1}}}

\title{\textit{Don’t Overthink It}: A Survey of Efficient R1-style Large Reasoning Models}

\author{%
Linan Yue$^{1,2}$, Yichao Du$^{4}$, Yizhi Wang$^{1,2}$, 
Weibo Gao$^{3}$, Fangzhou Yao$^{3}$, \\\textbf{Li Wang}$^{4}$, \textbf{Ye Liu}$^{3}$, \textbf{Ziyu Xu}$^{1,2}$, \textbf{Qi Liu}$^{3}$,  \textbf{Shimin Di}$^{1,2}$, \textbf{Min-Ling Zhang}$^{1,2}$\thanks{Corresponding Author}
  \\
  1: School of Computer Science and Engineering, Southeast University \\
  2:Key Laboratory of Computer Network and Information Integration (Southeast University), \\Ministry of Education\\
  3: University of Science and Technology of China \& State Key Laboratory of Cognitive Intelligence\\4: Alibaba Group\\
  \texttt{
    \{lnyue,wang\_yz,zyxu,shimin.di,zhangml\}@seu.edu.cn;  
  }\\
    \texttt{
    \{ycdu666,yeliu.liuyeah\}@gmail.com;
  }\\
  \texttt{
    \{weibogao,fangzhouyao,wl063\}@mail.ustc.edu.cn,
     qiliuql@ustc.edu.cn
  }\\
\faGithub \,\, \texttt{Github:\url{https://github.com/yuelinan/Awesome-Efficient-R1-style-LRMs}}
}

\begin{document}

\maketitle
\begin{figure}[h]
  \centering
   \vspace{-0.5cm}
  \includegraphics[width=11.cm]{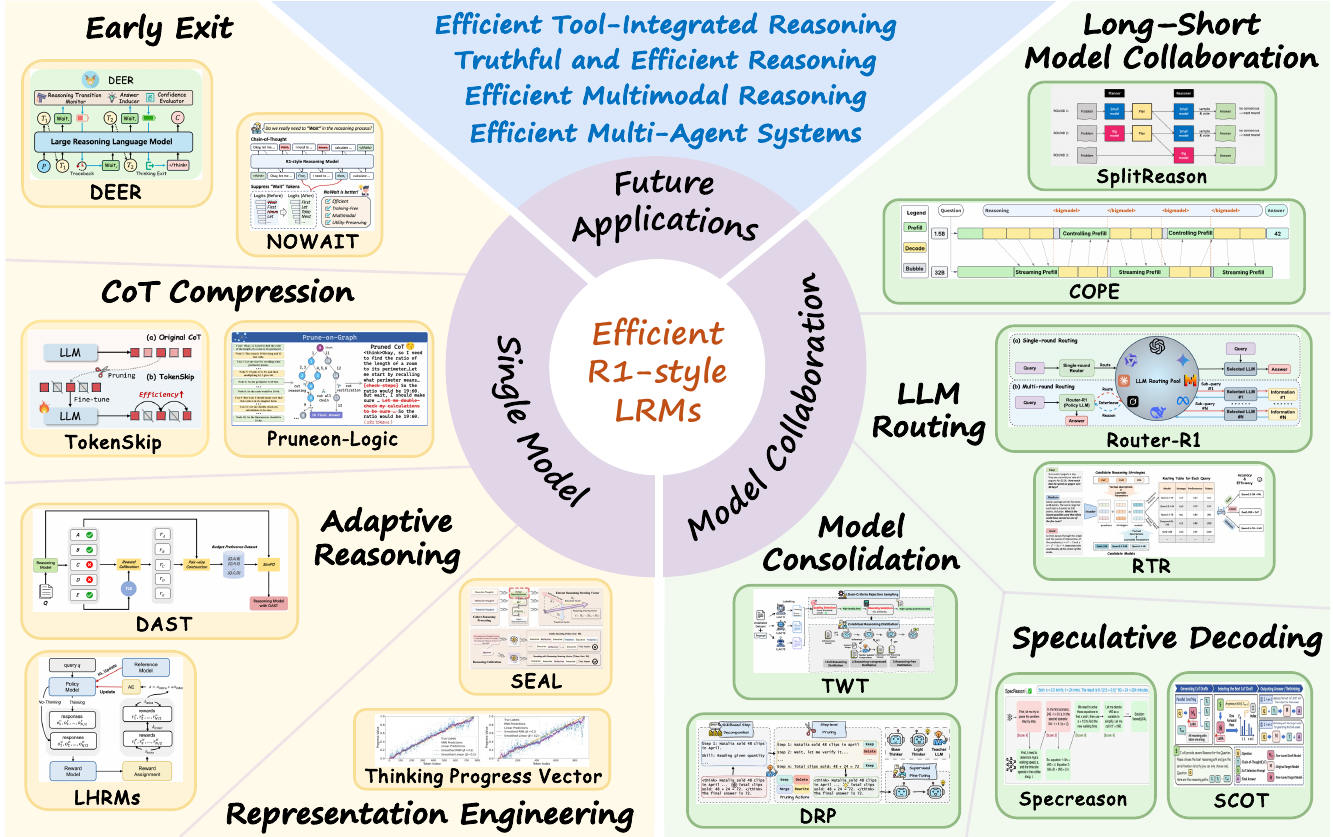}
  \caption{
     Taxonomy, Representative Methods and Future Applications of Efficient R1-style LRMs.
  } 
  \label{taxonomy1}
\end{figure}



\begin{abstract}
 Recently, Large Reasoning Models (LRMs) have gradually become a research hotspot due to their outstanding performance in handling complex tasks. Among them, DeepSeek R1 has garnered significant attention for its exceptional performance and open-source nature, driving advancements in the research of R1-style LRMs. Unlike traditional Large Language Models (LLMs), these models enhance logical deduction and decision-making capabilities during reasoning by incorporating mechanisms such as long chain-of-thought and self-reflection through reinforcement learning. However, with the widespread application of these models, the problem of \textbf{\textit{overthinking}} has gradually emerged. Specifically, when generating answers, these models often construct excessively long reasoning chains with redundant or repetitive steps, which leads to reduced reasoning efficiency and may affect the accuracy of the final answer.
To this end, various efficient reasoning methods have been proposed, aiming to reduce the length of reasoning paths without compromising model performance and reasoning capability. By  reviewing the current research advancements in the field of efficient reasoning methods systematically, we categorize existing works into two main directions based on the lens of single-model optimization versus model collaboration: (1) Efficient Reasoning with Single Model, which focuses on improving the reasoning efficiency of individual models; and (2) Efficient Reasoning with Model Collaboration, which explores optimizing reasoning paths through collaboration among multiple models.
Besides,  we maintain a public GitHub repository that tracks the latest progress in efficient reasoning methods.
We hope this survey not only consolidates recent advances but also introduces a novel organizational framework for understanding efficient reasoning, framing it through the lens of single-model optimization versus model collaboration.


\end{abstract}

\section{Introduction}
In recent years, Large Language Models (LLMs) have made groundbreaking progress in natural language processing tasks. However, when dealing with complex tasks like mathematical reasoning, multi-hop question answering, and program verification, LLMs still fall short in their reasoning abilities. As a result, Large Reasoning Models (LRMs) have attracted increasing attention \citep{xu2025towards,li2025system,chen2025towards}. These models enhance structured reasoning and advanced cognitive abilities by introduceing Long Chain-of-Thought (Long CoT) and self-reflection methods, enabling them to tackle complex problems more effectively. Representative works include OpenAI o1 \citep{jaech2024openai}, DeepSeek R1 \citep{guo2025deepseek}, Kimi~1.5 \citep{team2025kimi}, and QwQ \citep{qwq}.
In particular, DeepSeek R1 has become a benchmark for R1-style LRMs due to its outstanding reasoning accuracy and open-source accessibility, where the reasoning paths are commonly marked by the <think> and </think> tags.

With the widespread deployment of R1-style LRMs in practical applications, the issue of ``\textit{overthinking}'' has gradually emerged \citep{chen2024not,team2025kimi}. Specifically, when generating answers, the model often constructs lengthy CoT, sometimes introducing redundant or ineffective intermediate reasoning steps. This not only significantly reduces reasoning efficiency and increases computational costs, but the extra thinking may also lead to increased uncertainty and variance in the output, thereby affecting the accuracy of the final result \citep{suvra2025does}. For example, when handling a math problem that could be solved in three steps, the model might generate a redundant reasoning process with more than twenty steps, ultimately degrading overall performance. Furthermore, overthinking may introduce security risks, increasing the likelihood of the model being vulnerable to malicious attacks \citep{kuo2025h,fang2025safemlrm}. As a result, enabling models to ``\textit{think less but more accurately}'' has become a critical challenge in current reasoning model research.

\begin{figure}[t] 
\centering 
\scalebox{0.50}{
\begin{forest}
    for tree={ 
    font=\small, 
    semithick, rounded corners=2pt,
    text width=4cm, font=\bfseries,
    grow=0, 
    child anchor=west, 
    parent anchor=east, 
    anchor=west, 
    align=center,
    calign=center, 
    s sep=30pt, 
    l sep=14pt, 
    draw,
    inner sep=2.5mm,
    text centered,
    edge path={ 
      \noexpand\path[\forestoption{edge}]
      (!u.parent anchor) -- +(5pt,0) |- (.child anchor)\forestoption{edge label};
    }}
  [\rotatebox{90}{Efficient R1-style LRMs}, 
  fill=purple!10,align=center, text width=0.6cm, s sep=20pt, 
    [Future Applications (\secref{section5}), 
    fill=blue!3,
      for children={fill=blue!3,text width=20.3cm,align=left}, 
      [(1) Efficient Multimodal Reasoning (\secref{section5.1}); (2) Efficient Tool-Integrated Reasoning  (\secref{section5.2}); (3)  Efficient Multi-Agent Systems  (\secref{section5.3}); \\(4) Truthful and Efficient Reasoning (\secref{section5.4}) \,\,\,\,\,\,\,\,\,\,\,\,\,\,\,\,\,\,\,\,\,\,\,\,\,\,\,\,\,\,\,\,\,\,\,\,\,\,\,\,\,\,\,\,\,\,\,\,\,\,\,\,\,\,\,\,\,\,\,\,\,\,\,\,\,\,\,\,\,\,\,\,\,\,\,\,\,\,\,\,\,\,\,\,\,\,\,\,\,\,\,\,\,\,\,\,\,\,\,\,\,\,\,\,\,\,\,\,\,\,\,\,\,\,\,\,\,\,\,\,\,\,\,\,\,\,\,\,\,\,\,\,\,\,\,\,\,\,\,\,\,\,\,\,\,\,\,\,\,\,\,\,\,\,\,\,\,\,\,\,\,\,\,\,\,\,\,\,\,\,\,\,\,\,\,\,\,\,\,\,\,\,\,\,\,\,\,\,\,\,\,\,\,\,\,\,\,\,\,\,\,\,\,\,\,\,\,\,\,\,\,\,\,\,\,\,\,\,\,\,\,\,\,\,\,\,\,\,\,\,\,\,\,\,\,\,\,\,\,\,\,\,\,\,\,\,\,\,\,\,\,\,\,\,\,\,\,\,\,\,\,\,\,\,\,\,\,\,\,\,\,\,\,\,\,\,\,\,\,\,\,\,\,\,\,\,\,\,\,\,\,\,\,\,\,\,\,\,\,\,\,\,\,\,\,\,\,\,\,\,\,\,\,\,\,\,\,\,\,\,\,\,\,\,]
    ]
        [Efficient Reasoning \\with Model Collaboration \\(\secref{section4}), 
        fill=green!3,
      for children={fill=green!3, align=center}, 
      [Speculative Decoding (\secref{speculative}), for children={fill=green!3, draw, align=left,  text width=15.3cm}, [e.g. RSD \citep{liao2025reward}; SpecRouter \citep{wu2025specrouter}; SpecReason \citep{pan2025specreason}; \\Speculative Thinking \citep{yang2025speculative}; SCoT \citep{wang2025efficient};\,\,\,\,\,\,\,\,\,\,\,\,\,\,\,\,\,\,\,\,\,\,\,\,\,\,\,\,\,\,\,\,\,\,\,\,\,\,\,\,\,\,\,\,\,\,\,\,\,\,\,\,\,\,\,\,\,\,\,\,\,\,\,\,\,\,\,\,\,\,\,\,\,\,\,\,\,\,\,\,\,\,\,\,\,\,\,\,\,\,\,\,\,\,\,\,\,\,\,\,\,\,\,\,\,\,\,\,\,\,\,\,\,\,\,\,\,\,\,\,\,\,\,\,\,\,\,\,\,\,\,\,\,\,\,\,\,\,\,\,]]
      [Model Consolidation (\secref{konwledge}), for children={fill=green!3, draw, align=left,  text width=15.3cm}, [e.g. TwT \citep{xu2025twt}; DAR \citep{wu2025concise}; DRP \citep{jiang2025drp}; \\ReCUT \citep{jin2025recut}; \,\,\,\,\,\,\,\,\,\,\,\,\,\,\,\,\,\,\,\,\, \,\,\,\,\,\,\,\,\,\,\,\,\,\,\,\,\,\,\,\,\, \,\,\,\,\,\,\,\,\,\,\,\,\,\,\,\,\,\,\,\,\, \,\,\,\,\,\,\,\,\,\,\,\,\,\,\,\,\,\,\,\,\, \,\,\,\,\,\,\,\,\,\,\,\,\,\,\,\,\,\,\,\,\,\,\,\,\,\,\,\,\,\,\,\,\,\,\,\,\,\,\,\,\,\, \,\,\,\,\,\,\,\,\,\,\,\,\,\,\,\,\,\,\,\,\, \,\,\,\,\,\,\,\,\,\,\,\,\,\,\,\,\,\,\,\,\, \,\,\,\,\,\,\,\,\,\,\,\,\,\,\,\,\,\,\,\,\, \,\,\,\,\,\,\,\,\,\,\,\,\,\,\,\,\,\,\,\,]]
      [LLM Routing (\secref{route}), for children={fill=green!3, draw, align=left,  text width=15.3cm}, [e.g. RouteLLM \citep{ong2024routellm}; GraphRouter \citep{feng2024graphrouter}; IRT-Router \citep{song2025irt}; \\RouterBench \citep{hu2024routerbench}; R2R \citep{fu2025r2r}; Router-R1 \citep{zhang2025router}; \\TagRouter \citep{chen2025tagrouterlearningroutellms};   RTR \citep{pan2025route};\,\,\,\,\,\,\,\,\,\,\,\,\,\,\,\,\,\,\,\,\,\,\,\,\,\,\,\,\,\,\,\,\,\,\,\,\,\,\,\,\,\,\,\,\,\,\,\,\,\,\,]]
      [Long–Short \\Model Collaboration (\secref{long-short-model}), for children={fill=green!3, draw, align=left,  text width=15.3cm}, [e.g. SplitReason \citep{akhauri2025splitreason}; ThoughtMani \citep{liu2025thought}; CoThink \citep{fan2025cothink}; \\PLAN-AND-BUDGET \citep{lin2025plan}; VeriThinker \citep{chen2025verithinker};\\ FoReaL-Decoding \citep{li2025makes}; COPE \citep{lee2025collaborative}; ThinkSwitcher \citep{liang2025thinkswitcher};]]
    ]
    [Efficient Reasoning \\with Single Model (\secref{single}), fill=yellow!4,
      for children={fill=yellow!4}, 
     [Representation\\Engineering (\secref{Repe}), for children={fill=yellow!4, draw,align=left,  text width=15.3cm},[e.g. SEAL \citep{chen2025seal}; Pre-allocated Direction Vectors \citep{sheng2025reasoning};  \\Thinking Progress
Vector \citep{eisenstadt2025overclocking}; Manifold Steering \citep{huang2025mitigating};\,\,\,\,\,\,\,\,\,\,\,\,\,\,\,\,\,\,\,\,\, \,\,\,\,\,\,\,\,\,\,\,\,\,\,\,\,\,\,\,\,\, \,\,\,\,\,\,\,\,\,\,\,\,\,\,\,\,\,\,\,\,\, \,\,\,\,\,\,\,\,\,\,\,\,\,\,\,\,\,\,\,\,\, \,\,\,\,\,\,\,\,\,\,\,\,\,\,\,\,\,\,\,\,\,\,\,\,\,\,\,\,\,\,\,]]
      [Adaptive Reasoning (\secref{adaptive}), for children={fill=yellow!4, draw, align=left,  text width=15.3cm},[e.g. LHRMs \citep{jiang2025think}; Ada-R1 \citep{luo2025adar1}; DAST \citep{shen2025dast};  \\Guided by Gu \citep{ghasemabadi2025guided}; Thinker \citep{chung2025thinker}; CAR \citep{lu2025prolonged}; \\AdaCtrl \citep{huang2025adactrl}; Thinkless \citep{fang2025thinkless}; AdaCoT \citep{lou2025adacot};\\ ACPO \citep{cheng2025incentivizing}; ARM \citep{wu2025arm};  LASER \citep{liu2025learn}; \\HAPO \citep{huang2025hapo}; SOL \citep{yi2025shorterbetter}; ALP \citep{xiang2025just};\\ SelfBudgeter \citep{li2025selfbudgeter};\,\,\,\,\,\,\,\,\,\,\,\,\,\,\,\,\,\,\,\,\, \,\,\,\,\,\,\,\,\,\,\,\,\,\,\,\,\,\,\,\,\, \,\,\,\,\,\,\,\,\,\,\,\,\,\,\,\,\,\,\,\,\, \,\,\,\,\,\,\,\,\,\,\,\,\,\,\,\,\,\,\,\,\, \,\,\,\,\,\,\,\,\,\,\,\,\,\,\,\,\,\,\,\,\,\,\,\,\,\,\,\,\,\,\,\,\,\,\,\,\,\,\,\,\,\,\,\,\,\,\,\,\,\,\,\, \,\,\,\,\,\,\,\,\,\,\,\,\,\,\,\,\,\,\,\,\, \,\,\,\,\,\,\,\,\,\,\,\,\,\,\,\,\,\,\,\,\, \,\,\,\,\,\,\,\,\,\,\,\,\,\,\,\,\,\,\,\,\, \,\,\,\,\,\,\,\,\,\,\,\,\,\,\,\,\,\,\,\,\,\,\,\,\,\,\,\,\,\,\,\,\,\,\,\,\,\,\,\,\,\,\,\,\,\,\,\,\,\,\,\, \,\,\,\,\,\,\,\,\,\,\,\,\,\,\,\,\,\,\,\,\, \,\,\,\,\,\,\,\,\,\,\,\,\,\,\,\,\,\,\,\,\, \,\,\,\,\,\,\,\,\,\,\,\,\,\,\,\,\,\,\,\,\, \,\,\,\,\,\,\,\,\,\,\,\,\,\,\,\,\,\,\,\,\,\,\,\,\,\,\,\,\,\,\,]]
      [CoT Compression (\secref{cot-compress}), for children={fill=yellow!4, draw, align=left,  text width=15.3cm,},[e.g. CTS \citep{yuan2025not}; TokenSkip \citep{xia2025tokenskip}; RPC \citep{song2025reasoning}; \\Adaptive GoGI-Skip \citep{zhuang2025accelerating}; PIR \citep{xiao2025limopro}; SPIRIT \citep{cui2025stepwise}; \\DTO \citep{an2025don}; Prune-on-Logic \citep{zhao2025can}; $A^{*}$-Thought \citep{xu2025thought}; \\LS-Mixture SFT \citep{yu2025long}; AutoL2S \citep{luo2025autol2s};  LC-R1 \citep{cheng2025optimizing};\,\,\,\,\,\,\,\,\,\,\,\,\,\,\,\,\,\,\,\,\, \,\,\,\,\,\,\,\,\,\,\,\,\,\,\,\,\,\,\,\,\, \,\,\,\,\,\,\,\,\,\,\,\,\,\,\,\,\,\,\,\,\, \,\,\,\,\,\,\,\,\,\,\,\,\,\,\,\,\,\,\,\,\, \,\,\,\,\,\,\,\,\,\,\,\,\,\,\,\,\,\,\,\,\,\,\,\,\,\,\,\,\,\,\,]]
      [Early Exit (\secref{early}), for children={fill=yellow!4, draw, align=left,  text width=15.3cm}, 
      [e.g. DEER \citep{yang2025dynamic};  CONCISE \citep{qiao2025concise}; Elastic \citep{xu2025scalable}; \\Flashthink \citep{jiang2025flashthink};
       NOWAIT \citep{wang2025wait}; TIP \citep{wang2025thoughts};\\
       S-GRPO \citep{dai2025s};\,\,\,\,\,\,\,\,\,\,\,\,\,\,\,\,\,\,\,\,\, \,\,\,\,\,\,\,\,\,\,\,\,\,\,\,\,\,\,\,\,\, \,\,\,\,\,\,\,\,\,\,\,\,\,\,\,\,\,\,\,\,\, \,\,\,\,\,\,\,\,\,\,\,\,\,\,\,\,\,\,\,\,\, \,\,\,\,\,\,\,\,\,\,\,\,\,\,\,\,\,\,\,\,\,\,\,\,\,\,\,\,\,\,\,\,\,\,\,\,\,\,\,\,\,\,\,\,\,\,\,\,\,\,\,\, \,\,\,\,\,\,\,\,\,\,\,\,\,\,\,\,\,\,\,\,\, \,\,\,\,\,\,\,\,\,\,\,\,\,\,\,\,\,\,\,\,\, \,\,\,\,\,\,\,\,\,\,\,\,\,\,\,\,\,\,\,\,\, \,\,\,\,\,\,\,\,\,\,\,\,\,\,\,\,\,\,\,\,\,\,\,\,\,\,\,\,\,\,\,\,\,\,\,\,\,\,\,\,\,\,\,\,\,\,\,\,\,\,\,\, \,\,\,\,\,\,\,\,\,\,\,\,\,\,\,\,\,\,\,\,\, \,\,\,\,\,\,\,\,\,\,\,\,\,\,\,\,\,\,\,\,\, \,\,\,\,\,\,\,\,\,\,\,\,\,\,\,\,\,\,\,\,\, \,\,\,\,\,\,\,\,\,\,\,\,\,\,\,\,\,\,\,\,\,\,\,\,\,\,\,\,\,\,\,]
      ]
    ]
  ]
\end{forest}}
\caption{Taxonomy of efficient R1-style LRMs and future applications.} 
\label{fig:efficient_reasoning_taxonomy} 
\end{figure}
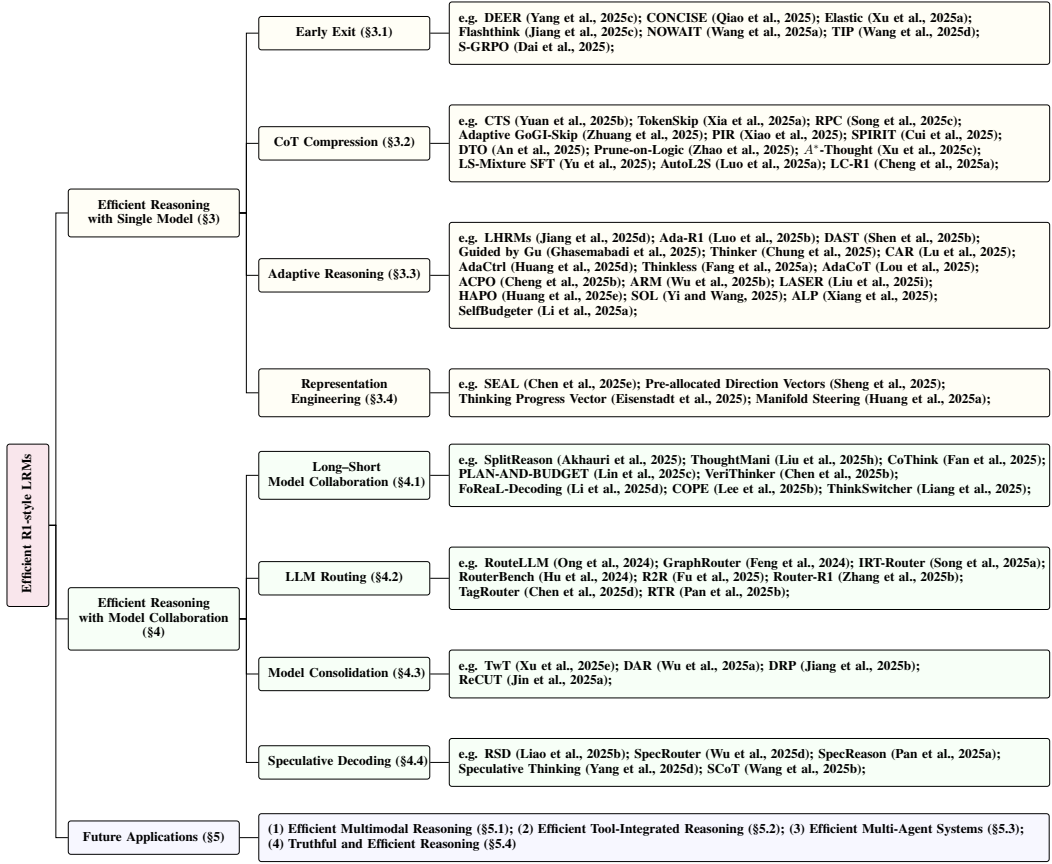

To this end, recent studies have explored methods to improve reasoning efficiency from multiple dimensions, leading to several preliminary survey studies. As shown in Table~\ref{tab:survey_comparison}, these studies \citep{liu2025efficient,qu2025survey,feng2025efficient,sui2025stop,wang2025harnessing,xu2025towards} mostly focus on training process, explicit and implicit CoT for effective reasoning. However, in contrast to previous works, in this survey, we present a new categorization perspective based on the lens of single-model optimization versus model collaboration, systematically reviewing cutting-edge research from 2025 onward. As shown in Figure \ref{taxonomy1} and Figure \ref{fig:efficient_reasoning_taxonomy}, we categorize existing efficient reasoning methods into two main directions: 

\textit{(1) Efficient Reasoning with Single Model}, which focuses on optimizing the reasoning path within a single model to improve computational efficiency. Specific strategies include Early Exit, CoT Compression, Adaptive Reasoning, and Representation Engineering (RepE) based Efficient Reasoning.

\textit{(2) Efficient Reasoning with Model Collaboration}, which focuses on enhancing the reasoning~efficiency through collaborative methods among multiple models. Related methods include Long–Short Model Collaboration, LLM Routing, Model Consolidation, and Speculative Decoding.

\begin{table}[htbp]
    \centering
    \small 
    \caption{Comparison of Existing Survey Papers}
    \renewcommand{\arraystretch}{0.9} 
    \resizebox{\textwidth}{!}{ 
        \begin{tabular}{lccccccccc} 
            \toprule 
            \textbf{Survey Paper} 
            & \makecell{\textbf{Focus on} \\ \textbf{Overthinking}}
            & \makecell{\textbf{Frontier} \\ \textbf{RepE Methods}}
            & \makecell{\textbf{Frontier} \\ \textbf{Model Collaboration}} 
            & \textbf{Taxonomy}  \\
            \midrule
         
            \citet{liu2025efficient}  & \checkmark  & $\times$ & $\times$ & Explicit/Implicit CoT  \\ 
            \citet{qu2025survey}    & \checkmark  & $\times$ & $\times$ & Training Process \\ 
            \citet{feng2025efficient}     & $\times$  & $\times$ & $\times$ & Short CoT/Small Model/Fast Decoding \\ 
            \citet{sui2025stop}          & \checkmark   & $\times$ & $\times$ & Training Process\\ 
            \citet{wang2025harnessing} &$\times$ &$\times$&\checkmark&Post-training/Test-time\\
            \citet{xu2025towards} & $\times$ &$\times$ &$\times$ &Reinforced Reasoning\\
            \midrule
            \textbf{Ours}   & \checkmark  & \checkmark & \checkmark & \makecell{ Single-model Optimization \textit{Vs.} \\Model Collaboration} \\ 
            \bottomrule 
        \end{tabular}
    }
    \label{tab:survey_comparison}
    
\end{table}

The framework of this survey is summarized as follows:

\begin{itemize}[leftmargin=*]
\item Section \ref{section2} introduces LRMs and the overthinking problem they face during reasoning, as well as the goals of efficient reasoning.

\item Section \ref{single} introduces efficient reasoning with a single model, exploring how to optimize a single model’s reasoning process to enhance efficiency.

\item Section \ref{section4} discusses efficient reasoning through model collaboration, focusing on how collaborative mechanisms among multiple models can improve reasoning efficiency.

\item Section \ref{section5} looks ahead to future development applications, covering frontier fields such as multimodal efficient reasoning, tool-integrated reasoning, multi-agent systems and truthful reasoning.
\end{itemize}

\section{Preliminaries}
\label{section2}
\subsection{Large Reasoning Models}

The OpenAI proposed o1 model \citep{jaech2024openai} has sparked widespread interest in LRMs. For example, \cite{qi2024mutual} introduce rStar, a self-play based mutual reasoning mechanism that significantly enhances the reasoning capabilities of small language models (SLMs) without relying on model fine-tuning or guidance from more powerful models. \cite{zhang2024rest} propose a tree search reinforced self-training method guided by process rewards, which automatically generates high-quality reasoning paths through tree search, effectively improving the model’s coherence and reasoning performance. Marco-o1 \citep{zhao2024marco} employs self-play and Monte Carlo Tree Search (MCTS) to generate long CoT data with reflection and error correction abilities. During inference, MCTS and process rewards jointly guide the model to explore an improved reasoning space, yielding higher-quality answers. These methods typically emphasize modeling process reward mechanisms and using MCTS in test-time scaling.

With the release of DeepSeek-R1 \citep{guo2025deepseek}, researchers have increasingly focused on constructing R1-style LRMs. Such models rely solely on rule-based reward functions, such as accuracy and format  rewards, for reinforcement learning (RL) training, which effectively unlocks long CoT reasoning capabilities and exhibits certain reflective  behaviors. In this section, we provide the following definition for R1-style LRMs:

Given any input question $x$, the output of an R1-style LRM consists of two parts: (1) a reasoning process $c = \{c_1, c_2, \dots, c_t\}$ composed of multiple reasoning units; and (2) a final answer $y$. Here, $c_i$ denotes the $i$-th reasoning step or segment. Some reasoning units contain specific key tokens (e.g., ``\textit{wait}'' and ``\textit{alternatively}''), which often signal the model’s ``\textit{Aha moment}'', reflecting the reflective transitions in the reasoning. In practice, to obtain $c_i$, the delimiter ``$\backslash n$$\backslash n$'' is commonly used to separate reasoning units.

\subsection{Overthinking Problem}
Overthinking refers to the tendency of LRMs to generate unnecessarily long, redundant, or overly complex reasoning paths during task execution, which can lead to response latency, increased computational cost, and even degraded answer accuracy \citep{chen2024not,team2025kimi,cuadron2025danger}.
In R1-style LRMs, overthinking typically manifests in the following ways:
\begin{enumerate}[leftmargin=*]
  \item \textbf{Overthinking Simple Problems:} In real-world applications, R1-style LRMs often generate detailed and complete CoT for all inputs, even for simple queries such as  ``\textit{What is 2 + 3?}''. 
  \item \textbf{Unconfident Reasoning Behavior:} During reasoning, LRMs often engage in self-verification and reflection. However, when deciding whether to reflect, the model may exhibit low confidence in its intermediate outputs, leading to unnecessary repeated reflection and \textit{self-doubt} style reasoning loops, thereby exacerbating the overthinking issue \citep{chen2025verithinker}.
\end{enumerate}

To mitigate such issues, recent studies have focused on \textit{efficient reasoning}, which aims to reduce the length and latency of reasoning paths while preserving answer accuracy and reflective behavior.

\section{Efficient Reasoning with Single Model}
\label{single}
Efficient reasoning with single model aims to achieve efficient reasoning by optimizing the reasoning process of a single model. This approach focuses on minimizing computational resources and reasoning time while maintaining reasoning accuracy, ensuring that the model can quickly and accurately generate answers. Specific methods include Early Exit (section \ref{early}), CoT Compression (section \ref{cot-compress}), Adaptive Reasoning (section \ref{adaptive}), and Representation Engineering-based Efficient Reasoning (section \ref{Repe}).

\subsection{Early Exit}
\label{early}
Early Exit in reasoning refers to the mechanism by which a LLM dynamically determines whether it has acquired sufficient information during the reasoning process, and then terminates generation before completing the full CoT, making the final prediction based solely on the current reasoning content. Interestingly, studies have shown that even when reasoning is terminated prematurely, the model’s prediction performance can often match that of full CoT reasoning \citep{liao2025fractured}. As such, Early Exit has emerged as a key research direction for enhancing the efficiency of R1-style reasoning models.

The core challenge of Early Exit lies in determining when a model should stop thinking. Existing approaches primarily address this question from three perspectives:
\begin{enumerate}[leftmargin=*]
  \item \textbf{Monitoring-based Early Exit:} These methods aim to dynamically monitor the model's internal reasoning state to decide whether reasoning should be terminated.
  \item \textbf{Generation Control-based Early Exit:} These methods manipulate the model’s generation behavior directly, e.g., by detecting and modifying the logits of specific trigger tokens, thereby preventing the model from producing redundant content.
  \item \textbf{Adaptive Early Exit:} These approaches allow the model to autonomously decide when to stop reasoning, without relying on pre-defined monitors or trigger tokens.
\end{enumerate}

\subsubsection{Monitoring-based Early Exit}
These methods continuously monitor the model’s internal states or generated content to dynamically assess whether the current reasoning process is sufficient, thereby determining whether to terminate the generation of the reasoning chain early \citep{zhu2025think}. Based on the type of monitoring signal utilized, these approaches can be further categorized into four subtypes:

\textbf{\textit{(1) Confidence-based termination.}}
This method relies on the model's confidence in its current reasoning state to decide whether to stop. When the model exhibits high confidence in an intermediate result, it can stop the reasoning process early and directly output the current answer.
Specifically, \cite{yang2025dynamic} propose a training-free dynamic early-exit method called DEER. This method identifies pivotal tokens (e.g., ``\textit{wait}'') within long CoT sequences and replaces them with guiding tokens such as ``\textit{final answer}'' to prompt the LLM to produce a tentative answer based on the current reasoning. The confidence of this answer is then evaluated. If it exceeds a predefined threshold, it is directly output, otherwise, the model rolls back to the turning point and continues reasoning.
Similarly, \cite{qiao2025concise} identify two typical redundancy patterns in reasoning: \textit{Confidence Deficit}, where the model underestimates the validity of its correct intermediate steps and engages in unnecessary reflection, and \textit{Termination Delay}, where the model continues reasoning even after generating a correct answer. To address these issues, \cite{qiao2025concise} propose the CONCISE framework. It first introduces a confidence injection technique that inserts high-confidence phrases into the reasoning path to enhance trust in intermediate steps. Then, an early stopping module with a confidence detector monitors the model's confidence level and halts generation once it exceeds a defined threshold.

\textbf{\textit{(2) Entropy-based dynamic control method.}}
Unlike approaches that rely on explicit confidence signals, entropy-based control methods adopt an information-theoretic perspective \citep{shannon1948mathematical}, focusing on the trend of information gain throughout the reasoning process to determine whether reasoning should be terminated. Specifically, \cite{yong2025think} first introduce two metrics: InfoBias and InfoGain. Then, they empirically find that longer reasoning paths tend to exhibit higher information bias and diminishing information gain, especially when generating incorrect answers. Based on these findings, the authors propose an entropy-based reasoning mechanism, where the reasoning process is automatically terminated if the InfoGain falls below a preset threshold for $k$ consecutive steps. Additionally, an entropy regularization term is incorporated during training to encourage the model to terminate reasoning early when InfoGain becomes minimal.

\textbf{\textit{(3) Budget-constrained early termination method
.}}
This method explicitly imposes a token usage budget on the reasoning process \citep{xu2025scalable,li2025selfbudgeter,liu2025answer}, forcing termination when the consumption approaches or reaches the upper limit, thereby controlling computational cost. For example, \cite{xu2025scalable} propose the elastic reasoning method, which divides the token budget into two parts: one for the thinking stage and one for the answering stage. When the thinking-stage budget is exhausted, reasoning is forcibly terminated to ensure sufficient budget remains for answer generation.
\cite{liu2025answer} further propose a supervised learning method that leverages internal model activation \citep{3740645,liu-etal-2024-enhancing-language} sequences. An LSTM-based reasoning progress estimator is trained to dynamically predict the optimal stopping point based on model activation patterns, allowing for timely and effective early termination.

\textbf{\textit{(4) Probe-based early termination method
.}}
This method does not rely on explicit confidence scores or entropy signals. Instead, it utilizes external probe models or verification mechanisms to predict the correctness of intermediate reasoning results and decide whether to terminate generation early \citep{zhu2025think,zhang2025reasoning,jiang2025flashthink}. Specifically, \cite{zhang2025reasoning} segment the full reasoning process into multiple chunks, and at the end of each chunk, the model generates an intermediate answer, which is labeled as either correct or incorrect (using a binary supervision signal $y$). The final hidden state of each chunk is extracted as the input feature $x$, forming a training set of $(x, y)$ pairs. Based on this, a multilayer perceptron (MLP) probe is trained to predict the probability that the current answer is correct. During inference, if the predicted probability exceeds a certain threshold, reasoning is terminated early and the current answer is output.
Similarly, \cite{jiang2025flashthink} also segment the reasoning content into multiple fragments and employs a pre-trained verification model to assess whether the current fragment contains sufficient information to arrive at the correct answer. If so, the reasoning process is terminated; otherwise, it continues.

\subsubsection{Generation Control-based Early Exit}
This category of methods bypasses internal state monitoring and instead intervenes directly in the decoding process to compress reasoning paths and improve efficiency \citep{wang2025wait,wang2025thoughts,liu2025efficientreasoningsuppressionselfaffirmation}. For example, \cite{wang2025wait} propose the NOWAIT method, which employs a logit processor during decoding to explicitly prohibit the generation of specific tokens that trigger unnecessary reflection. For any predefined token, the corresponding logit value is assigned a large negative value, effectively suppressing the sampling of such tokens and enabling more efficient reasoning. 
Similarly, \cite{wang2025thoughts} employ the Thought Switching Penalty (TIP), which adjusts the predicted logits of tokens associated with reasoning branch transitions, further reducing unnecessary digressions in the reasoning trajectory.
Common reflection or switching-related tokens include:

\begin{figure}[h]
  \centering
  \scalebox{0.88}{
  \begin{tcolorbox}
    “wait”, “alternatively”, “hmm”, “but”, “however”, “alternative”, “another”, “check”, “double-check”, “oh”, “maybe”, “verify”, “other”, “again”, “now”, “ah”, “any”.

  \end{tcolorbox}}
  \label{case_study}
  \vspace{-0.2cm}
\end{figure}

Additionally, \cite{liu2025answer} propose another generation control strategy. Unlike the aforementioned suppression approaches, their goal is to enhance the generation probability of the end-of-thinking token (i.e., the </think> token) to encourage early stopping. Specifically, \cite{liu2025answer} introduce an adaptive probability enhancement method for the </think> token. During decoding, the authors apply a linear logit boosting strategy to increase the relative competitiveness of the </think> token, making it more likely to be sampled when the model’s output distribution is concentrated, thereby achieving early termination.

\subsubsection{Adaptive Early Exit}

This category of methods does not rely on explicit monitoring signals or specific tokens and decoding control. Instead, it introduces learned policies that enable models to autonomously determine when they have thought enough, thereby achieving adaptive early stopping of the reasoning path. To this end, \cite{dai2025s} propose a reinforcement learning approach named S-GRPO (Serial-Group Decaying-Reward Policy Optimization). This method inserts  \textit{early exit} instructions at different positions within a single reasoning chain to construct multiple serial reasoning path groups. It then applies a decaying reward strategy based on the exit position: the earlier the model terminates reasoning while still producing a correct answer, the higher the reward it receives. This guides the model to stop reasoning as early as possible without sacrificing accuracy. Compared with the parallel-path-based GRPO method \citep{shao2024deepseekmath}, S-GRPO models reasoning sufficiency in a more fine-grained manner, improving both reasoning efficiency and answer accuracy.

\subsection{CoT Compression}
\label{cot-compress}
Chain-of-Thought Compression (CoT Compression) methods aim to shorten the reasoning chains of LLMs while preserving their original reasoning effectiveness, thereby improving inference efficiency and deployment feasibility. A straightforward approach is to leverage prompt learning to guide models to autonomously generate more concise reasoning paths \citep{renze2024benefits,nayab2024concise}. For instance, \cite{han2024token} propose the prompt ``\textit{Use at most $k$ tokens}'' to explicitly constrain reasoning length. \cite{aytes2025sketch} further introduce the Sketch-of-Thought (SoT) framework, which employs structured prompts to elicit clear and concise reasoning steps. This method incorporates three reasoning paradigms (i.e., conceptual chaining, chunked symbolism, and expert lexicons) to adapt to different reasoning tasks, and integrates a lightweight routing model for dynamic paradigm selection. While effective, these approaches often rely on manually crafted prompts and lack adaptability, limiting their applicability across diverse tasks.
To improve the generality and efficiency of CoT compression, existing research mainly follows three perspectives:

\begin{enumerate}[leftmargin=*]
  \item \textbf{Granularity-based CoT Compression:} These methods compress CoTs at different granularities, including token-level, step/chunk-level, and chain-level.
  \item \textbf{Parallel thinking-based Compression:} By sampling multiple reasoning paths from the model, these methods compare and aggregate the paths to construct a compressed version.
  \item \textbf{Reward-based Compression:} Instead of directly pruning reasoning paths, these methods design compression reward functions that encourage the model to learn adaptive compression strategies during training.
\end{enumerate}

\subsubsection{Granularity-based CoT Compression}
These methods build upon existing CoT reasoning paths $c$ to generate compressed data tuples $(x, \hat{c}, y)$, where $x$ denotes the model input, $y$ is the final answer, and $\hat{c}$ represents the compressed reasoning path. Based on such compressed datasets, LLMs are further fine-tuned via Supervised Fine-tuning (SFT) to achieve effective reasoning chain compression. Depending on the granularity focus of CoT compression, these methods can be further categorized into the following three types:

\textbf{\textit{(1) Token-level compression based on importance estimation.}}
These methods focus on estimating the importance of individual tokens within the reasoning chain and removing less important tokens for compression \citep{yuan2025not,xia2025tokenskip,song2025reasoning,zhuang2025accelerating,lee2025well}. Specifically, \cite{yuan2025not} propose Conditional Token Selection (CTS), which trains a reference model to assess token-level importance during reasoning, and dynamically removes redundant tokens using metrics such as \textit{perplexity} to construct a compressed dataset. \cite{xia2025tokenskip} further introduce TokenSkip, which estimates token importance and applies a compression threshold to retain only high-weight tokens, yielding a concise version of the reasoning chain.
Reasoning Path Compression (RPC) \citep{song2025reasoning} improves inference efficiency by periodically compressing the key-value (KV) cache in LRMs. The method uses attention mechanisms to score recently generated tokens by their importance, retaining only high-impact entries to reduce redundant computation.

Besides, to ensure coherence in the compressed reasoning process, \cite{zhuang2025accelerating} propose the Adaptive GoGI-Skip method. This approach first quantifies the contribution of each token to the final prediction by computing the loss. Then, it introduces a dynamic pruning strategy based on uncertainty: when the model’s prediction entropy is high, indicating greater task difficulty, pruning is reduced; conversely, when entropy is low, more aggressive pruning is allowed. In addition, an Adaptive N-Constraint mechanism is used to limit the number of consecutively pruned tokens based on the moving average of entropy, preserving the continuity of reasoning. Based on these strategies, a compressed dataset is constructed for retraining the model.

\textbf{\textit{(2) Step-level compression based on importance estimation.}}
Unlike token-level compression methods, this category of approaches partitions the reasoning chain into higher-level semantic units, such as steps, chunks, or segments, and performs selection at that granularity \citep{xiao2025limopro,cui2025stepwise,wang2025r1,an2025don,zhao2025can,xu2025thought,lin2025trimr}. Compared with token-level methods, these approaches place greater emphasis on semantic coherence and logical completeness.
Specifically, \cite{xiao2025limopro} propose the Perplexity-based Importance Refinement (PIR) framework, which systematically categorizes reasoning steps into progressive and functional types. By leveraging perplexity-based scoring, PIR selectively removes low-importance functional steps and constructs a refined dataset for model fine-tuning to improve inference efficiency.
Similarly, \cite{cui2025stepwise} introduce the SPIRIT algorithm, which addresses both few-shot CoT prompting and fine-tuning scenarios. SPIRIT iteratively removes or merges reasoning steps based on perplexity, while designing demonstration refinement or training data optimization strategies to ensure that the resulting reasoning chains remain both concise and semantically coherent.
\cite{wang2025r1} divide the model-generated solution into well-structured semantic chunks and generate multiple simplified candidates for each chunk. A greedy search is then conducted across chunks to select the candidate that best balances conciseness and fidelity, measured by low language model loss.
\cite{an2025don} propose the Dynamic Thought Optimization (DTO) framework, which partitions the reasoning chain into segments representing different cognitive modes. DTO evaluates these segments to selectively reinforce beneficial ones and prune detrimental ones, constructing preference pairs to perform preference learning.

In addition, some studies move beyond the traditional linear CoT structure by converting the reasoning process into more structured representations such as graphs. For instance, \cite{zhao2025can} introduce the Prune-on-Logic framework, which transforms CoT into a logical graph and prunes redundant or ineffective nodes to achieve structurally consistent compression with stronger logical validity.
\cite{xu2025thought} propose $A^{*}$-Thought, which models the reasoning process as a search tree. This method employs bidirectional importance estimation (via bidirectional language modeling) and leverages $A^{*}$ search to optimize reasoning paths, effectively compressing long chains and accelerating LLMs inference.

\textbf{\textit{(3) Chain-level compression via rewriting.}}
This line of research focuses on rewriting the entire CoT to reduce its overall length and complexity \citep{yu2025long,luo2025autol2s}. Unlike token- or step-level pruning strategies, chain-level methods offer a global perspective, aiming to simplify the reasoning process holistically in terms of semantics and structure.
Specifically, \cite{yu2025long} propose the LS-Mixture SFT approach, which rewrites long CoT sequences into more concise versions while preserving their reasoning structure. These rewritten short chains are then mixed with the original long-chain data for supervised fine-tuning, effectively reducing redundant reasoning behavior in the model.
Similarly, \cite{luo2025autol2s} introduce the Auto Long-Short Reasoning (AutoL2S) method. They construct training data that includes both long and short CoT paths, where the short CoTs are rewrote with a special <EASY> token at the beginning to indicate the corresponding problem is simple. The model is then fine-tuned on this mixed dataset. After training, if the model generates the <EASY> token during inference, it follows a simplified reasoning path, enabling dynamic compression of the reasoning process.

\subsubsection{Parallel thinking–based Compression}
Unlike previous approaches that optimize a single sampled reasoning path, this category is inspired by Best-of-N (BoN) sampling strategies \citep{beirami2024theoretical,amini2024variational,agarwal2025first}, which parallelize the generation of multiple candidate reasoning paths and select the superior ones to guide compression \citep{munkhbat2025self,suvra2025does}.
Specifically, \cite{munkhbat2025self} leverage self-generated reasoning paths and combine naive BoN sampling, few-shot prompting (FS), and few-shot guided BoN (FS-BoN) strategies to identify the shortest correct reasoning path. This path is then used to construct a compressed dataset for SFT, enabling efficient reasoning compression.
Similarly, \cite{suvra2025does} also propose a BoN-style sampling strategy for efficient reasoning. Rather than explicitly shortening a single reasoning path, they evenly allocate the total token budget across $N$ parallel paths and use parallel decoding to simultaneously generate multiple candidate chains. The best-performing path is selected as the final output.

Beyond path selection, some studies explore parallel execution mechanisms to reduce reasoning time. Specifically, \cite{biju2025sprint} propose the SPRINT framework, which consists of a planner and multiple executors. During reasoning, the planner generates multiple subplans from the reasoning context, which are then executed in parallel by independent agents to accelerate inference. \cite{hassid2025don} further suggest a strategy where $k$ reasoning paths are generated in parallel, and once the shortest $m$ of them ($k \geq m$) are completed, the generation of the remaining paths is terminated. The answers from the $m$ finished paths are then aggregated via majority voting to select the final reasoning outcome.

\subsubsection{Reward-based Compression}
This category of methods \citep{cheng2025optimizing,zeng2025done} does not directly prune or rewrite reasoning paths. Instead, it introduces compression reward mechanisms to guide models in autonomously learning compression strategies through reinforcement learning, thereby enabling dynamic optimization of reasoning content. Specifically, \cite{cheng2025optimizing} first propose two key principles: brevity and sufficiency. Guided by these principles, they design the LC-R1 post-training method based on GRPO \citep{shao2024deepseekmath}. This approach incorporates a compression reward focused on the </think> token, encouraging the model to terminate reasoning promptly after generating the correct answer and premature termination before completing effective reasoning is penalized to prevent excessive compression from harming prediction accuracy. Through this mechanism, the model adaptively balances compression rate and accuracy.
Additionally, \cite{zeng2025done} combine chain rewriting with this approach by reconstructing the original long-form CoT paths into structured multi-turn interactive processes. Specifically, \cite{zeng2025done} first convert raw CoT into a multi-turn dialogue format to build training data, which is then initialized by SFT. Subsequently, reinforcement learning using GRPO \citep{shao2024deepseekmath} is applied, with the reward design including interaction rounds as an optimization target, encouraging the model to complete accurate reasoning in fewer turns, thereby compressing the overall reasoning process.

Notably, the above methods are all single-model compression approaches. For multi-model collaborative compression mechanisms, please refer to Section \ref{model-dill}.

\subsection{Adaptive Reasoning}
\label{adaptive}
Adaptive Reasoning aims to enable LLMs to dynamically adjust the depth and length of their reasoning processes based on task requirements and input complexity. Unlike conventional methods that rely on static reasoning paths, adaptive reasoning empowers models with the ability to ``\textit{decide whether to reason, how long to reason, and how to reason}'' autonomously.

To achieve this goal, adaptive reasoning methods typically integrate RL frameworks, where carefully designed reward mechanisms guide the model to learn optimal reasoning strategies under varying conditions. Existing research in this area can be broadly categorized into three main perspectives:

\begin{enumerate}[leftmargin=*]
  \item \textbf{RL-based Adaptive Reasoning:} Inspired by DeepSeek R1 \citep{guo2025deepseek}, these methods focus on reward design, by encouraging the model to learn when and how to reason effectively.
  \item \textbf{Reasoning-mode Switching:} These methods emphasize the decision of whether to reason or which reasoning mode to choose. The core idea is to assess the complexity of a given input and dynamically select an appropriate reasoning strategy, such as direct answering, short reasoning, or in-depth reasoning.
  \item \textbf{Adaptive Reasoning with Length Reward:} As an extension of RL-based methods, these methods explicitly target reasoning path length. Models are guided to learn what constitutes an optimal reasoning length by setting length reward objectives.
\end{enumerate}

\subsubsection{RL-based Adaptive Reasoning}
\label{rl-based}
This class of methods incorporates RL frameworks and carefully designed reward functions to guide LLMs in dynamically adjusting their reasoning process based on input complexity. As a core approach to adaptive reasoning, RL-based methods are also widely employed in the subsequent sections on reasoning-mode switching and adaptive reasoning with length reward. Here, we focus on representative works that model adaptive reasoning primarily through reinforcement learning. Based on whether a ``\textit{Warm-up}'' phase is introduced prior to RL training, existing methods can be further categorized into the following two types:

\textbf{\textit{(1) RL methods with a Warm-up phase.}}
These methods typically begin with a SFT phase using mixed reasoning-path data (i.e., both short and long chains), which enables the model to acquire the ability to perform diverse reasoning strategies \citep{jiang2025think,wang2025adaptive}. This warm-up stage is followed by an RL phase to further optimize the model’s adaptive decision-making ability. For instance, \cite{jiang2025think} propose Large Hybrid Reasoning Models (LHRMs). They first fine-tune the model with a combination of long-chain and short-chain reasoning samples, equipping it with both reasoning styles. Then, they introduce Hybrid Group Policy Optimization to train the model to adaptively choose between reasoning modes. An evaluation metric called Hybrid Accuracy is also proposed to measure the model’s effectiveness in selecting the appropriate reasoning strategy.
Similarly, \cite{wang2025adaptive} also perform SFT using a mix of short and long CoT samples, followed by RL. Their reward design incorporates intra-group accuracy to guide reasoning mode selection and a first-token logits loss to optimize initial decoding behavior.

Distinct from the above, \cite{luo2025adar1} propose Ada-R1, a two-stage adaptive reasoning framework. In the first stage, they merge the parameters of a long-chain reasoning model and a standard LLM to form a unified model capable of generating both long and short reasoning paths. The second stage introduces a dual-level optimization mechanism: group-level preference optimization guides the model to select short or long reasoning modes based on input characteristics, while instance-level preference encourages the model to generate more concise reasoning under the constraint of maintaining accuracy, thereby improving overall reasoning efficiency.

\textbf{\textit{(2) RL methods without a Warm-up phase.}}
Unlike the previous approaches, this class of methods \citep{shen2025dast,ghasemabadi2025guided,chung2025thinker,yang2025think,qi2025optimizing} directly trains LLMs using RL without a supervised warm-up stage.
Specifically, DAST \cite{shen2025dast} builds an explicit mapping between problem difficulty and response length, introducing a metric called Token Length Budget (TLB). For each input query, multiple reasoning paths are sampled and their corresponding TLB values are calculated. Then, preference pairs are constructed based on reasoning quality and efficiency. These pairs are used to fine-tune the model via  SimPO \citep{meng2024simpo}, enabling it to learn adaptive reasoning strategies.
Guided by Gut (GG) \citep{ghasemabadi2025guided} leverages intrinsic signals from the LLM’s own generation process, such as token-level confidence, to guide the reasoning search, without relying on external verification models. Through RL, the model is trained to optimize its internal confidence estimation, and is coupled with a self-guided tree-search strategy. This framework significantly reduces computational costs while preserving reasoning quality.

In addition, \cite{chung2025thinker} propose Thinker, a four-stage reasoning framework guided by RL. The model learns to dynamically decide among four steps: Fast Thinking $\rightarrow$ Verification $\rightarrow$ Slow Thinking $\rightarrow$ Summary. Initially, the model performs fast thinking to produce a draft answer. If verification fails, it proceeds to a slow-thinking phase for in-depth correction. Finally, it summarizes the full reasoning path. Each stage is paired with a custom-designed reward function to enable adaptive reasoning across different reasoning demands.

\subsubsection{Reasoning-mode Switching}
This category of methods dynamically determines whether reasoning is necessary and which reasoning mode to adopt by assessing the complexity of the current input. Typical strategies involve switching between multiple modes such as fast/slow thinking or thinking/no-thinking \citep{zhang2025othink}. Based on how the switching mechanism is implemented, these methods can be further divided into two subcategories:

\textbf{\textit{(1) Token-based reasoning mode switching.}}
These approaches explicitly inject control tokens (e.g., <fast\_think> and <slow\_think>) to indicate different reasoning modes \citep{cheng2025incentivizing, huang2025adactrl, fang2025thinkless, tu2025learning}.
For example, \cite{cheng2025incentivizing} propose the Adaptive Cognition Policy Optimization (ACPO) framework, which introduces <fast\_think> and <slow\_think> tokens to enable dynamic switching between fast and slow thinking in LRMs. Concretely, they construct reasoning paths on a high-quality math dataset by prompting diverse outputs of varying lengths, and use GPT-4 to conduct fine-grained comparisons. Important reasoning steps are labeled as slow-thinking, while redundant or simple steps are tagged as fast-thinking. These mixed-mode paths are used to perform SFT, followed by RL using an online TLB reward \citep{shen2025dast} to guide adaptive depth control based on input difficulty.
Similarly, \cite{huang2025adactrl} introduce the AdaCtrl framework, which uses a cold-start SFT stage on a dataset labeled with special tokens like <Easy> and <Hard> to establish initial mode-switching ability. In the subsequent RL phase, a difficulty-aware response length reward and difficulty calibration mechanism are introduced to enhance adaptive reasoning across tasks.
Thinkless \citep{fang2025thinkless} utilizes the first token (<think> or <short>) in the output sequence to control reasoning behavior. A Decoupled Group-wise Relative Policy Optimization (DeGRPO) algorithm is then used to jointly optimize both mode selection and final answer accuracy.

Beyond explicit control tokens, other works have explored implicit switching signals. For instance, \cite{tu2025learning} use ellipsis-style prompts (...) to invoke optional reasoning behavior in R1-style models. \cite{zhang2025adaptthink} guide models to switch between “Thinking” and “NoThinking” modes depending on problem complexity, interpreting an initial </think> token as a no-thinking decision. Similarly, \cite{lou2025adacot} propose the AdaCoT framework, which constructs two types of samples: one with ``<think>reasoning\_steps</think>answer'' for tasks requiring reasoning, and another with ``<think></think>answer'' for straightforward queries, training the model to control whether and when to reason. Finally, \cite{lu2025prolonged} introduce the Certainty-based Adaptive Reasoning (CAR) framework. Trained on mixed reasoning paths, the model initially generates concise answers and uses perplexity as a proxy for uncertainty. If confidence is low, a longer CoT response is triggered, enabling a dynamic trade-off between efficiency and performance.

\textbf{\textit{(2) Multi-mode reasoning switching.}}
In contrast to binary reasoning mode switching, some approaches \citep{wu2025arm,xie2025interleaved} further extend the diversity of reasoning paradigms by enabling the model to adaptively select among three or more reasoning strategies.
Specifically, \cite{wu2025arm} propose the Adaptive Reasoning Model (ARM), which supports four distinct reasoning formats. The model is trained in two stages: in the first stage, SFT is used to equip the model with multiple reasoning paradigms; in the second stage, an improved group-wise relative policy optimization algorithm (Ada-GRPO) is introduced to guide the model in dynamically selecting the optimal reasoning mode based on task requirements.
In a different vein, \cite{xie2025interleaved} introduce the Interleaved Reasoning framework. Unlike the traditional ``\textit{think-then-answer}'' linear paradigm, this method adopts an interleaved generation structure of ``\textit{thinking–answering–thinking}'', where intermediate informative answers are generated during the reasoning process. These answers serve as both guidance for subsequent steps and as verifiable reward signals, enabling the model to iteratively refine its reasoning and converge toward the correct final answer.

\subsubsection{Adaptive Reasoning with Length Reward}
This category of methods focuses on controlling the length of the generated reasoning paths, typically by introducing explicit reward shaping or penalty mechanisms to guide the model toward eliminating redundant content while preserving prediction accuracy \citep{gao2025far,li2025aalc,luo2025o1,hou2025thinkprune,su2025thinking,aggarwal2025l1,yuan2025efficient,song2025walk,liu2025bingo,ling2025fast}.
Among them, \cite{liu2025learn} propose LASER (Length-Aware Shaping via Reinforcement learning), a RL approach that designs a stepwise reward function based on target length. It also introduces a difficulty-aware dynamic reward scheme, balancing reasoning efficiency with task performance.
\cite{huang2025hapo} introduce History-Aware Policy Optimization (HAPO), which maintains a history of the shortest correct answer length. Responses shorter than this value are rewarded, while those exceeding it are penalized, even if correct.
\cite{yi2025shorterbetter} operate under the assumption that each question has a Sample Optimal Length (SOL). They obtain this SOL by sampling multiple candidates per input, identify the shortest correct one, and use it to guide reward assignment via GRPO. Adaptive Length Penalty (ALP) \citep{xiang2025just} performs multiple rollouts per input to estimate a solve rate (i.e., success ratio), and dynamically adjusts the length penalty: inputs with high solve rates are penalized more heavily to discourage overlong reasoning; those with low solve rates receive weaker penalties, allowing longer reasoning chains to ensure correctness.
Finally, from the perspective of token budget, \cite{li2025selfbudgeter} propose the SelfBudgeter framework. In the training phase, the model first undergoes cold-start fine-tuning to learn how to predict the required token budget before answering. Then, GRPO is used to further optimize this prediction process, encouraging the model to minimize token usage while strictly adhering to the predicted length budget without compromising accuracy.

\subsection{Representation Engineering based Efficient Reasoning}
\label{Repe}
Representation Engineering (RepE) \citep{zou2023representation} treats the internal representations of neural networks as fundamental units of operation, aiming to precisely control model behavior by analyzing and transferring these representations. In recent years, RepE has demonstrated broad applicability in domains such as hallucination mitigation \citep{li2023inference}, safety enhancement \citep{arditi2024refusal}, and reasoning capability improvement \citep{zhang2024uncovering,tang2025unlocking}.

These methods typically follow a two-stage pipeline of representation extraction and representation control. In the first stage, hidden representations from models under different states are collected, and directional vectors are computed by taking the difference between representations. These vectors capture key behavioral shifts. In the second stage, these vectors are injected into the hidden states of the target model to steer its behavior.
For example, in reasoning capability improvement scenarios, 
given a set of problems $X = \{x_1, x_2, ..., x_n\}$ and two models $M_{\text{short}}$ with short-chain reasoning and $M_{\text{long}}$ with long-chain reasoning, we can achieve the following steps:

In the representation extraction phase, the difference vectors can be computed as:
\begin{equation}
  \delta_i = M_{\text{long}}(x_i) - M_{\text{short}}(x_i).
\end{equation}
Aggregating these can yield a reasoning-mode steering vector:
\begin{equation}
p_L = \frac{1}{|X|} \sum_{i=1}^{|X|} \delta_i = \frac{1}{|X|} \sum_{i=1}^{|X|} \left( M_{\text{long}}(x_i) - M_{\text{short}}(x_i) \right).
\end{equation}
Then, in the representation control stage, given a target model $M_{\text{target}}$ and its input’s hidden state $M_{\text{long}}(x_{i})$, a controled hidden state is constructed as:
\begin{equation}
\tilde{M}_{\text{long}}(x_{i}) = M_{\text{long}}(x_{i}) + \lambda_p \cdot p_L,
\end{equation}
where $\lambda_p$ is a scaling hyperparameter. This intervention nudges the target model toward the reasoning style of $M_{\text{long}}$, effectively enhancing its reasoning depth.

In this section, we focus on the application of RepE for mitigating overthinking \citep{chen2025seal,sheng2025reasoning,eisenstadt2025overclocking,huang2025mitigating,ma2025cot,liu2025fractional,azizi2025activation,lin2025controlling}. Specifically, \citep{chen2025seal} propose SEAL (Steerable rEAsoning caLibration), a framework that categorizes reasoning units into execution, reflection, and transition, and constructs steering vectors to represent efficient reasoning directions. These vectors are injected into the hidden space during decoding to dynamically suppress redundant reflections and non-essential transitions, while preserving core execution logic.
\cite{sheng2025reasoning} show that the number of reasoning tokens can be predicted from input activations via a linear probe, indicating the model's implicit control over reasoning length. They construct Pre-allocated Direction Vectors, whose subtraction reduces reasoning depth and accuracy, while addition extends reasoning and improves performance.
Similarly, \cite{eisenstadt2025overclocking} find that LLMs implicitly track their reasoning progress via internal signals. Based on this, they propose the Thinking Progress Vector to enable fine-grained control over reasoning length, thus preventing overthinking. Differently, \cite{huang2025mitigating} conduct mechanistic interpretability analyses and find that overthinking behaviors lie on a specific low-dimensional manifold in the model’s activation space. They introduce Manifold Steering, which projects interventions onto this manifold to avoid high-dimensional noise, thereby reducing computational overhead and performance degradation caused by overthinking.

\section{Efficient Reasoning with Model Collaboration}
\label{section4}
Efficient reasoning with model collaboration aims to enhance reasoning efficiency and accuracy in LLMs by enabling cooperation between multiple LLMs, each leveraging distinct reasoning strengths. Unlike single model efficient reasoning method described in section \ref{single}, collaborative frameworks strategically combine long-chain reasoning models (long CoT) that excel at handling complex tasks and short-chain reasoning models (short CoT) that are lightweight and efficient for general tasks. This synergy allows for more fine-grained and cost-effective control of the reasoning process. Specific methods include Long–Short Model Collaboration (section \ref{long-short-model}), LLM Routing (section \ref{route}), Model Consolidation (section \ref{konwledge}), and Speculative Decoding (section \ref{speculative}).

\subsection{Long–Short Model Collaboration}
\label{long-short-model}
Long–Short Model Collaboration refers to approaches that integrate the complementary advantages of Long CoT and Short CoT models through dynamic interactions. This section focuses on ``\textit{two-model}'' setups, where one long CoT and one short CoT model are jointly involved in the reasoning process. Depending on which model plays the dominant role in the interaction, these methods can be categorized into three types:

\begin{enumerate}[leftmargin=*]
  \item \textbf{Short-to-Long Collaborative Reasoning:} These methods are short-model–centric, with the short CoT model handling most queries and selectively invoking the long CoT model for complex or uncertain reasoning tasks.
  \item \textbf{Long-to-Short Collaborative Reasoning:} In contrast, these methods are long-model–centric, where the long CoT model leads the reasoning and the short CoT model provides auxiliary support.
  \item \textbf{Long$\otimes$Short Interactive Reasoning:} These methods allow both models to interleave and alternate during the reasoning process, enabling multi-round interaction and dynamic control of reasoning depth and complexity.
\end{enumerate}

\subsubsection{Short-to-Long Collaborative Reasoning}
\label{s2l}
These methods typically designate the short CoT model as the primary reasoning agent or utilize it to plan the reasoning process, which then guides the long CoT model \citep{akhauri2025splitreason,liu2025thought,fan2025cothink,lin2025plan,chen2025verithinker,kim2025guiding}. Specifically, \citep{akhauri2025splitreason} propose a SplitReason framework where the short CoT model performs most reasoning steps while dynamically offloading complex substeps to the long CoT model. The approach enables collaborative reasoning between models by allowing the short model to delegate tasks it cannot handle. Training proceeds in two stages: first, a SFT phase teaches the short CoT model to insert offloading boundaries marked by special tokens <bigmodel>...</bigmodel>; second, a RL phase based on GRPO optimizes the offloading behavior using a reward function that jointly considers accuracy, formatting consistency, and offloading ratio to balance performance and efficiency.
ThoughtMani \citep{liu2025thought} employs a short CoT model to generate a CoT, which is injected as a prompt between <think> and </think> tokens of the long CoT model. This design allows the long model to directly read and leverage the short model’s reasoning trajectory, resulting in more efficient and targeted reasoning.
Similarly, CoThink \citep{fan2025cothink} adopts a two-stage framework in which a lightweight instruction model first generates a high-level solution plan, which is then used to guide the long CoT model through detailed reasoning.
PLAN-AND-BUDGET \citep{lin2025plan} proposes a budget-aware planning framework that dynamically allocates reasoning budgets based on task structure and uncertainty. The short CoT model first decomposes the original question into sub-problems and estimates the complexity of each sub-problem using confidence scores. A normalized token budget is then assigned to each sub-task. During inference, the long CoT model generates reasoning for each sub-problem within its token budget, and an aggregation module compiles the final answer.
VeriThinker \citep{chen2025verithinker} introduces a Supervised Verification Fine-Tuning (SVFT) approach, enabling the short CoT model to self-verify the correctness of its output. If the answer is deemed reliable, it is returned directly. Otherwise, the model triggers long CoT reasoning to produce a more robust response.

\subsubsection{Long-to-Short Collaborative Reasoning}
This category of methods typically places the long CoT model as the primary reasoning agent, or leverages it to guide the short CoT model in completing subsequent reasoning steps \citep{li2025makes,she2025hawkeye}. For example, \cite{li2025makes} propose FoReaL-Decoding, a framework in which a strong leading model  (long CoT)  first generates the initial tokens of a sentence to establish the reasoning direction and style. Then, a lightweight draft model (short CoT) continues the generation to complete the response. To prevent the leading model from oversteering or dominating the reasoning process, FoReaL-Decoding incorporates a stochastic gating mechanism that dynamically controls the frequency of intervention by the leading model, ensuring a balanced division of labor and effective collaboration between the two models.

\subsubsection{Long$\otimes$Short Interactive Reasoning}

These methods \citep{ning2025not,lee2025collaborative,liang2025thinkswitcher} explore interleaved or collaborative reasoning between long and short CoT models to improve inference efficiency. Specifically, \cite{ning2025not} first fine-tune LLMs using synthetic instruction data to separately acquire long and short style reasoning capabilities. Based on this, they design a multi-turn dialogue–based RL method, where rewards are defined over final answer correctness, format, and reasoning length. The long CoT model is encouraged to focus on generating key reasoning steps, while the short CoT model completes the rest with concise reasoning, thereby improving both performance and efficiency.
COPE \citep{lee2025collaborative} introduces a multi-stage plan-and-reasoning framework. In Stage 1, the short CoT model handles both planning and reasoning. In Stage 2, the long CoT model takes over planning, while the short CoT model continues reasoning. In Stage 3, the long CoT model fully dominates both planning and reasoning. After each stage, candidate answers are collected through sampling and voting. If no consensus is reached, the system proceeds to the next stage for deeper reasoning.
ThinkSwitcher \citep{liang2025thinkswitcher} proposes a lightweight mode-switching module that dynamically selects between long and short CoT models without retraining the base reasoning models. Given an input question, the switcher takes its representation as input and predicts the expected performance of long and short chain reasoning paths. During training, ThinkSwitcher adopts a multi-sample evaluation strategy to generate multiple responses per reasoning mode, and constructs continuous supervision signals based on empirical solve rates, thereby avoiding instability from binary labels. At inference time, the model selects the optimal reasoning path based on the switcher’s prediction.

\subsection{LLM Routing}
\label{route}
LLM routing aims to dynamically select the most suitable model(s) from a model pool for each input query, thereby significantly reducing computational cost while maintaining reasoning performance. The model pool typically consists of multiple pretrained models with varying scales. For instance, simple questions such as “What is 2 + 3?” can be routed to lightweight models (e.g., GPT-2) instead of invoking LLMs  (e.g., DeepSeek-R1) to avoid the overthinking problem, thereby improving overall inference efficiency. Existing studies have proposed a variety of routing mechanisms, which can be broadly categorized into the following two types:
\begin{enumerate}[leftmargin=*]
  \item \textbf{Single-Step Routing:} These methods perform a one-time evaluation of the input query before inference, routing it to a single most appropriate model to complete the task. It is characterized by simplicity and fast response time.
  \item \textbf{Multi-Step Routing:} These methods enable dynamic routing to multiple models during the inference process, allowing for collaborative reasoning. It typically decides in real time whether to involve a more powerful model based on the current reasoning state.
\end{enumerate}

\subsubsection{Single-Step Routing}
This line of work typically selects a single model for inference per query, offering simplicity in implementation and low latency in response \citep{lu2023routing,chen2024routerdc,ding2024hybrid,zhuang2024embedllm,zhang2025long,chen2025harnessing}. For instance, RouteLLM \citep{ong2024routellm} introduces four representative routing strategies: Similarity-weighted Ranking, Matrix Factorization, BERT Classifier, and Causal LLM Classifier, to enable dynamic selection and switching between small and large models.

To improve the precision of routing decisions, a growing body of research focuses on aligning model capabilities with query characteristics. Specifically, GraphRouter \citep{feng2024graphrouter} employs graph neural networks (GNNs) to model the complex interactions among queries, models, and tasks, thereby optimizing model selection. IRT-Router \citep{song2025irt} incorporates Item Response Theory (IRT) \citep{woodruff1996estimation,gao2021rcd} to capture latent relationships between LLM capabilities and query attributes, enabling more fine-grained adaptation.
Some methods further introduce similarity-based routing mechanisms. For example, RouterBench \citep{hu2024routerbench} and \cite{shnitzer2023large} propose K-nearest neighbor (KNN) routing strategies, selecting candidate models by measuring similarity between the current input and historical queries.
TagRouter \citep{chen2025tagrouterlearningroutellms} presents a training-free model routing approach. It consists of three key modules: a TagGenerator that produces semantically relevant tags for each query, a TagScorer that learns mappings from tags to model performance using existing data, and a TagDecider that determines the final routing path based on these mappings.

In addition, to better leverage prior samples and model capability information, \cite{he2025self} construct a labeled dataset to distinguish whether a query requires reasoning, based on problem difficulty. They train a reasoning-mode selector accordingly. During inference, a lightweight pre-reasoning stage is introduced to extract capability-aware embeddings from intermediate model representations. These embeddings are used to estimate whether the current model can directly generate a high-quality answer. If the query is deemed complex, the reasoning mode is activated to generate a complete CoT. Otherwise, a generic mode is used to produce a concise response, thus effectively avoiding over-reasoning on simple tasks.

\subsubsection{Multi-Step Routing}
This class of methods \citep{shao2025route,zhang2025router,fu2025r2r,pan2025route} allows routing to different models multiple times during the inference process, typically making dynamic decisions based on the current reasoning state to determine whether additional models should be involved. This enables a flexible trade-off between performance and computational cost. Specifically, R2-Reasoner \citep{shao2025route} proposes a RL based framework for collaborative multi-model reasoning. It consists of two key components: a Task Decomposer, which splits complex tasks into well-structured and logically ordered subtasks, and a Subtask Allocator, which dispatches each subtask to the most appropriate model in a heterogeneous model pool based on its difficulty and characteristics. The training process involves SFT on a constructed dataset for both modules, followed by staged RL to alternately optimize their parameters, thereby enabling efficient and adaptive reasoning routes.
Router-R1 \citep{zhang2025router} formulates the routing process as a sequential decision-making problem and designs the router itself as a reasoning-capable LM. This setup allows dynamic alternation between ``\textit{thinking}'' and ``\textit{routing}'' during task execution, enabling the system to coordinate multiple models to collaboratively complete complex reasoning tasks. The reward function integrates reasoning format consistency, answer correctness, and computational cost, guiding the model toward an effective balance between performance and resource consumption.
R2R (Roads to Rome) \citep{fu2025r2r} further introduces a fine-grained, token-level routing strategy. The system initially lets a small LLM take the lead in reasoning, but selectively invokes a LLM at critical junctures where ambiguity or reasoning divergence is likely to occur. By combining automatic annotation with a lightweight router, R2R significantly reduces overall computation while maintaining reasoning accuracy.

Distinct from the above methods, which mainly focus on model routing, Route-To-Reason (RTR) \citep{pan2025route} expands the routing target by jointly routing both models and reasoning strategies. During inference, RTR not only dynamically selects which LLM to invoke, but also routes the query to the most suitable reasoning strategy module (e.g., PAL \citep{gao2023pal} or CoD \citep{xu2025chain}), thus enabling a structured and strategy-driven reasoning process.

\subsection{Model Consolidation}
\label{konwledge}
This class of methods aims to combine the strengths of LLM and SLM models to construct a new model with efficient reasoning capabilities, thereby significantly reducing computational cost while maintaining strong reasoning performance. Existing approaches can be broadly categorized into two types:
\begin{enumerate}[leftmargin=*]
  \item \textbf{Model Distillation:} These methods typically adopt a large model as the teacher and transfer its reasoning ability to a smaller student model. By incorporating techniques such as long-CoT compression during the distillation process, the student model is equipped for efficient reasoning.
  \item \textbf{Model Merging:} These methods merge the parameters of long-CoT and short-CoT models to integrate their complementary reasoning styles and capabilities, resulting in a new model that supports efficient and effective reasoning.
\end{enumerate}

\subsubsection{Model Distillation}
\label{model-dill}
This class of methods typically leverages LLMs to generate high-quality CoT, constructs new training datasets, and performs SFT on SLM to enable efficient transfer of reasoning capabilities \citep{xu2025twt,wu2025concise,jiang2025drp,wen2025light}.
For example, TwT \citep{xu2025twt} proposes a reasoning path synthesis framework in which multiple teacher models collaboratively generate diverse candidate CoT paths. These are filtered based on quality and diversity metrics to construct a high-quality reasoning dataset. Building upon this, TwT introduces Habitual Reasoning Distillation, a three-stage process. Specifically, the student model first learns from the complete reasoning paths generated by the teachers. The teacher then compresses and optimizes the reasoning paths based on the student's performance, creating simplified data for continued training.
Finally, the student model is trained solely on final answers, thereby acquiring the ability to complete tasks without relying on explicit reasoning chains.
Similarly, \cite{wu2025concise} propose an efficient distillation approach based on Difficulty-Aware Prompting (DAR). In this method, a LLMs (e.g., DeepSeek-R1 \citep{guo2025deepseek}) rewrites CoT paths by adapting them to the difficulty of the input problem, automatically generating more concise and adaptive reasoning paths. These are used to build the LiteCoT dataset, enabling the student model to learn compressed yet effective reasoning patterns.

In contrast to the above approaches that rely on teacher-generated CoT paths, DRP \citep{jiang2025drp} begins with initial reasoning paths generated by the student model and applies pruning via a teacher model. Specifically, the teacher identifies and removes redundant or irrelevant steps in the paths, merges semantically repetitive content, and outputs more compact and logically coherent reasoning units, which are then used to supervise the student.
These methods can be viewed as multi-model compression approaches, in contrast to single-model compression methods (see Section~\ref{cot-compress}).

\subsubsection{Model Merging}
This class of methods builds new models with adaptive reasoning capabilities by merging the parameters of long-CoT and short-CoT models, thereby balancing reasoning effectiveness and inference efficiency \citep{wu2025unlocking,jin2025recut,luo2025adar1,team2025kimi}. Specifically, \cite{wu2025unlocking} systematically investigate various model merging strategies, including Average Merging \citep{wortsman2022model}, Task Arithmetic \citep{ilharco2022editing}, TIES-Merging \citep{yadav2023resolving}, DARE \citep{yu2024language}, AIM \citep{nobari2025activation}, LoRE-Merging \citep{liu2025lore}, Twin-Merging \citep{lu2024twin}, and Sens-Merging \citep{liu2025sens}. Experimental results demonstrate that model merging can significantly reduce inference length, by up to 55\% in average response length, while preserving output quality, validating its effectiveness in enhancing reasoning efficiency.
Similarly, \cite{hu2025beyond} propose a three-stage framework for constructing reasoning LMs. In the first stage, multiple expert models are trained using modular RL, each specializing in a distinct reasoning paradigm such as deduction, induction, or abduction. Each expert is optimized via a reward function that combines format and answer correctness. In the second stage, the expert models are merged into a unified model using weighted parameter fusion. In the third stage, the merged model undergoes further fine-tuning on domain-specific tasks such as mathematics and programming, resulting in notable improvements in overall reasoning ability. This framework offers a viable paradigm for building efficient reasoning models.

In addition, \cite{jin2025recut} introduce ReCUT, a method that generates multiple reasoning paths via diverse sampling and constructs preference pairs based on both reasoning accuracy and path length. Two sub-models are then trained using Direct Preference Optimization (DPO) \citep{rafailov2023direct}, each targeting a different optimization goal. The final model is obtained by merging the two sub-models, achieving a favorable balance between quality and efficiency. Ada-R1 \citep{luo2025adar1} similarly merges a long CoT model with a general LLM to build a model that can handle reasoning tasks of varying depths. On top of this, RL is further employed to enhance inference efficiency and stability, with detailed training strategies described in Section~\ref{rl-based}.

\subsection{Speculative Decoding}
\label{speculative}
Speculative Decoding is a recently proposed technique for accelerating LLM inference \citep{3693232,zhang2025learning,liu2025pearl,huang2025poss,liao2025reward,wang2025accelerating,xia-etal-2024-unlocking,wu2025specrouter}. The core idea is to let a SLM quickly draft a segment of candidate tokens, which are then verified in parallel by the LLM. Only if the LLM deems these tokens to be consistent with its own likely generation are they accepted. Otherwise, it re-generates the corresponding content. This ``\textit{draft-then-verify}'' strategy significantly reduces the number of sequential decoding steps required by the LLM, thereby improving efficiency while preserving generation quality.
For instance, Reward-Guided Speculative Decoding (RSD) \citep{liao2025reward} allows a lightweight model to propose candidate reasoning steps, which are then evaluated using a reward function. Only when necessary does it invoke the LLM for correction, achieving a more flexible trade-off between accuracy and computational cost.
SpecRouter \citep{wu2025specrouter} introduces a multi-stage speculative decoding framework that replaces traditional static draft-target model pairs. It dynamically selects the most appropriate draft model and intermediate verification path based on task complexity and system load, reducing rejection rates and optimizing decoding throughput.
These methods can be viewed as a special case of Short-to-Long Collaborative Reasoning in section~\ref{s2l}, where the SLM proposes and the LLM verifies, enabling faster yet reliable generation through inter-model collaboration.

Since \citep{xia-etal-2024-unlocking} have provided a comprehensive survey of speculative decoding methods, this section focuses on recent advancements that specifically target LRMs \citep{pan2025specreason,yang2025speculative,wang2025efficient}.
For example, SpecReason \citep{pan2025specreason} proposes a speculative reasoning framework that performs fine-grained, adaptive delegation: semantically simple and non-critical reasoning steps are handled by a lightweight model, while the stronger model (Long CoT) verifies the semantic validity of these steps. If verified, the reasoning proceeds,  otherwise, the stronger model takes over to revise or continue the reasoning process.
Similarly, Speculative Thinking \citep{yang2025speculative} dynamically identifies reflective, uncertain, or self-negating tokens in the draft generated by the SLM. The LLM selectively intervenes at these critical reasoning junctures, enhancing the quality of reasoning for complex tasks while preserving overall efficiency.
SCoT (Speculative Chain-of-Thought) \citep{wang2025efficient} introduces a training-free speculative reasoning framework that generates multiple CoT drafts using a SLM and lets the LLM either select the most promising one or perform re-reasoning when needed. Unlike token-level speculative decoding, SCoT operates at the CoT-segment level, leveraging the SLM’s generation efficiency while reducing latency. It also employs LoRA-based \citep{hu2022lora} alignment between the draft and selector models to mitigate variance and redundancy in the generated drafts. This method exemplifies a lightweight collaborative paradigm for accelerating reasoning in LRMs.

\section{Future Applications}
\label{section5}
\subsection{Efficient Multimodal Reasoning}
\label{section5.1}
Multimodal reasoning models aim to tackle complex tasks that involve the integration of heterogeneous data sources such as text, images, and audio, and have attracted increasing attention in recent years \citep{li2025perception}. Among them, R1-style multimodal reasoning models have achieved notable performance improvements on complex reasoning tasks by introducing reinforcement learning mechanisms, particularly through the widespread adoption of the GRPO algorithm \citep{meng2025mm,yang2025r1,huang2025vision,shen2025vlm}.
Despite their success, these methods also reveal a more severe overthinking problem, characterized by redundant reasoning paths and repetitive reflective processes, which lead to substantial computational overhead. To address this challenge, a natural direction is to transfer efficient reasoning methods developed in the textual domain to multimodal scenarios \citep{lu2025prolonged}. However, there remains a lack of systematic evaluation and empirical analysis to assess the applicability and effectiveness of the various efficient reasoning methods summarized in this survey under multimodal settings. There is an urgent need to construct dedicated benchmark tasks for efficient multimodal reasoning, to evaluate the transferability, generalization, and practical benefits of these methods.

Furthermore, compared to textual reasoning, multimodal reasoning involves a more intricate ``\textit{perception–understanding–reasoning}'' pipeline, encompassing subtasks such as vision-language alignment and region grounding. When these components are entangled within a single reasoning path, they tend to introduce unnecessary computation and information noise. A more efficient strategy is to structurally decompose the multimodal reasoning process by clearly delineating the roles and boundaries of each stage. For example, Rex-Thinker \citep{jiang2025rex} divides the reasoning process into three stages: planning, action and summarization. Similarity, Visionary-R1 \citep{xia2025visionary} adopts a caption–reason–answer framework that first generates detailed image descriptions, followed by reasoning and answer generation. Building on such structured decomposition, future research may further explore stage-wise modeling and dynamic control of reasoning length, by applying stage-specific length rewards based on task complexity, thereby improving overall efficiency and stability without sacrificing reasoning quality.

\subsection{Efficient Tool-Integrated Reasoning}
\label{section5.2}
Tool learning aims to overcome the inherent limitations of LLMs in computation, memory, and access to external knowledge \citep{qu2025tool}. In recent years, it has been widely adopted to integrate external tools, such as code interpreters, calculators, and search engines, thereby enhancing the model’s adaptability and problem-solving capabilities. Existing Tool-Integrated Reasoning (TIR) methods primarily rely on SFT using reasoning paths extracted from stronger LLMs. While this approach can improve tool usage to some extent, it often restricts the model’s ability to explore tool invocation strategies autonomously, leading to rigid patterns and limited discovery of optimal solutions.
To enhance adaptability in tool use, recent research has introduced RL-based approaches that enable models to dynamically decide whether and which tools to invoke based on task requirements \citep{jin2025search,song2025r1,qian2025toolrl,peng2025learning}. For example, \cite{song2025r1} annotate reasoning paths in R1-style frameworks using <think> tags for internal thoughts, and introduce structured tags such as <begin\_of\_query>...<end\_of\_query> and <begin\_of\_documents>...<end\_of\_documents> in search scenarios to explicitly distinguish between query intents and retrieved results, thereby facilitating clearer modeling of the reasoning process.

However, even with such explicit annotation schemes, these TIR methods still suffer from the overthinking problem: models may excessively invoke external tools, resulting in unnecessary computational overhead and latency. In Retrieval-Augmented Generation (RAG) settings, the situation is further exacerbated when retrieved documents are noisy, prompting models to repeatedly reason over redundant or irrelevant content.
To mitigate these issues, future research could explore the following two directions: (1) incorporating reward mechanisms that penalize excessive tool calling, encouraging models to minimize redundant calls while maintaining answer accuracy; and (2) performing document refinement and filtering prior to the reasoning stage, removing uninformative or low-density content to reduce unnecessary inference costs at the source \citep{shi2025search}. These approaches hold promise for achieving a better efficiency–performance trade-off in TIR.

\subsection{Efficient Multi-Agent Systems}
\label{section5.3}
In multi-agent systems, multiple agents are typically required to collaborate on complex tasks, a process that heavily relies on efficient information exchange and strategic coordination \citep{zhang2024cut,li2025adaptive,wang2025agentdropout}. However, when individual agents suffer from overthinking, it can significantly slow down the overall system response and lead to substantial resource waste, ultimately degrading task execution efficiency at the system level.
To alleviate this issue, LLM Routing \citep{yue2025masrouter} has emerged as a promising solution. In this paradigm, the router serves as a central component of the agent architecture, dynamically assigning tasks to appropriate models to optimize resource allocation. Specifically, the router leverages task complexity, contextual cues, or historical interaction data to route simpler tasks to lightweight models and delegate more complex ones to powerful LRMs. This approach not only reduces the average computational cost but also improves system-wide responsiveness while maintaining the quality of task completion.
Furthermore, future research could explore agent-level reasoning budget scheduling, incorporating techniques such as confidence estimation and adaptive task analysis to enable more fine-grained and intelligent coordination across agents. These directions hold promise for building more efficient multi-agent reasoning systems.

\subsection{Truthful and Efficient Reasoning}
\label{section5.4}
Although R1-style LRMs demonstrate strong reasoning performance, their trustworthiness remains a significant challenge due to issues such as low safety \citep{kuo2025h,wang2025safety} and the generation of hallucinated information \citep{Hallucination,sun2025detection}. Existing efficient reasoning methods often overlook these trustworthiness risks during optimization. For example, CoT compression methods improve reasoning efficiency by shortening original long CoT sequences. However, they may inadvertently inherit and even amplify security vulnerabilities or hallucination problems present in LRMs. Therefore, ensuring model trustworthiness while enhancing reasoning efficiency is an important and urgent research direction for future work on efficient reasoning.

Furthermore, beyond conventional efficient reasoning evaluation metrics focused on accuracy, computational cost, and token usage, it is essential to develop methods to evaluate the trustworthiness of the reasoning process and results. Investigating the trade-offs between trustworthiness and accuracy also represents a promising direction for future research.

\section{Conclusion}
This paper presents a comprehensive survey of efficient reasoning, targeting the overthinking phenomenon commonly observed in R1-style Large Reasoning Models (LRMs). We propose a novel taxonomy that categorizes existing approaches into two major paradigms: single-model and multi-model reasoning. Furthermore, we outline several promising applications that stand to benefit from efficient reasoning, shedding light on potential extensions and new frontiers for future research. We hope this survey provides valuable insights and stimulates further work toward developing reasoning models that are not only capable, but also resource-efficient.

\bibliographystyle{ACM-Reference-Format}
\bibliography{ref}
\end{document}